\documentclass{article}
\usepackage{ijcai23}

\usepackage{times}
\usepackage{soul}
\usepackage{url}
\usepackage[hidelinks]{hyperref}
\usepackage[utf8]{inputenc}
\usepackage[small]{caption}
\usepackage{graphicx}
\usepackage{amsmath}
\usepackage{amsthm}
\usepackage{booktabs}
\usepackage{algorithm}
\usepackage{algorithmic}
\usepackage[switch]{lineno}

\usepackage{amsfonts}
\usepackage{subcaption}
\usepackage{bm}
\usepackage{bbm}
\usepackage{multirow}
\DeclareMathOperator*{\argmin}{arg\,min}

\usepackage{csquotes}
\usepackage{natbib}
\usepackage{cleveref}

\title{Ordinal Regression for Difficulty Estimation of StepMania Levels}

\author{
Billy Joe Franks$^{1}$\thanks{equal contribution}
\and
Benjamin Dinkelman$^{1*}$\and
Sophie Fellenz$^1$\And
Marius Kloft$^1$
\affiliations
$^1$ML group, University of Kaiserslautern-Landau, Germany\\
\emails
\{franks, b\_dinkelma19, fellenz, kloft\}@cs.uni-kl.de,
}

\begin{document}

\maketitle

\begin{abstract}
StepMania is a popular open-source clone of a rhythm-based video game. As is common in popular games, there is a large number of community-designed levels. It is often difficult for players and level authors to determine the difficulty level of such community contributions. In this work, we formalize and analyze the difficulty prediction task on StepMania levels as an ordinal regression (OR) task. We standardize a more extensive and diverse selection of this data resulting in five data sets, two of which are extensions of previous work. We evaluate many competitive OR and non-OR models, demonstrating that neural network-based models significantly outperform the state of the art and that StepMania-level data makes for an excellent test bed for deep OR models. We conclude with a user experiment showing our trained models' superiority over human labeling.
\end{abstract}

\section{Introduction}
\begin{figure}[tb]
\centering
    \includegraphics[width=\linewidth]{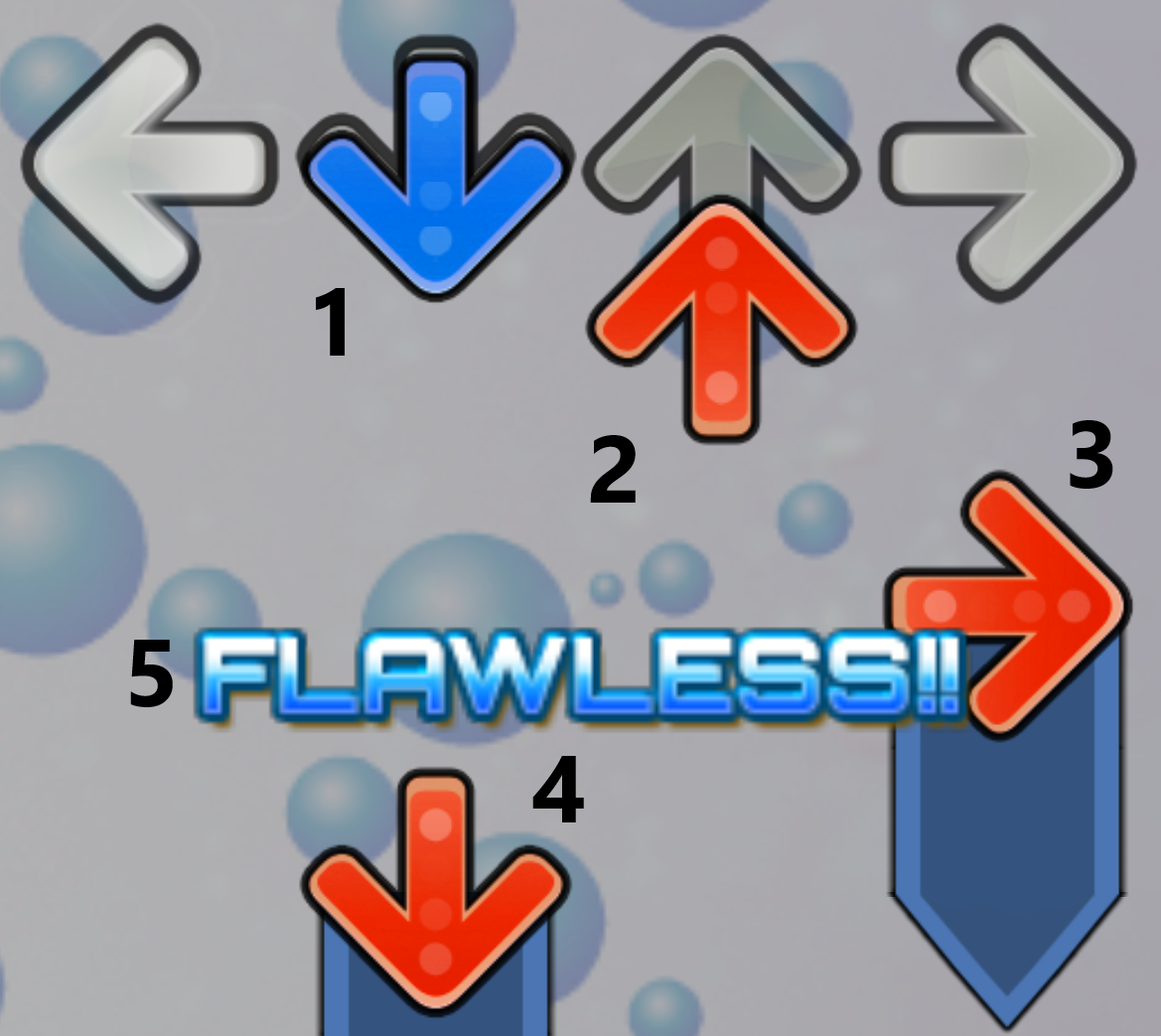}
    \caption{This example of StepMania play shows the basic steps tap (no. 1 and 2) and hold (no. 3 and 4). Steps that coincide with notes of certain levels, like a quarter (no. 2-4) or an eight (no. 1), have unique colors. Different steps can also be combined, requiring the player to hit two or more keys simultaneously. Depending on how accurately in time a step is hit, the player receives feedback (no. 5).}
    \label{fig:StepMania}
\end{figure}

Video game designers commonly order game levels in ascending order of difficulty. The first levels act as tutorials, while the later levels challenge the players and teach them new skills. However, games that rely heavily on community contributions lack communication present in game studios, leading to a more haphazard design and inconsistent game-level difficulties. Portal 2, Super Mario Maker, Happy Wheels, and Roblox are examples that profit heavily from community-created game levels. In this work, we focus on StepMania, a rhythm-based video game in which players step onto a keypad on the floor to the rhythm of a song. A level is represented by a sequence of directional inputs that must be hit at a specific time (see Fig. \ref{fig:StepMania}).

As difficulties are commonly represented as natural numbers (or ordinals), estimating the difficulty of video game levels is a natural ordinal regression (OR) task. OR has a long history dating back to at least the 18th century \citep{jian1727college}. It is commonly applied in the social sciences for modeling human preferences, as it allows for the representation of ordinal relationships. More recently, \citet{armstrong1989ordinal} raised greater interest from a machine learning (ML) perspective, while the use of deep learning started with its advent around the 2000s \citep{agarwal2001supporting}. Outside of the social sciences, OR also has applications in computer vision and natural language processing, which commonly require deep learning. Examples include age estimation \citep{niu2016ordinal}, sentiment analysis \citep{saad2019twitter}, or depth estimation \citep{fu2018deep}. However, it has not yet been applied to the difficulty prediction of video game levels.

There is a huge potential for applying ML methods such as OR on video games other than StepMania, but so far, the number of publicly available labeled data sets is limited.
Crawling the largest repository of StepMania data, StepMania Online\footnote{\url{https://search.stepmaniaonline.net/}}, results in 602 GB of labeled data. More data can also be found on various platforms frequented by avid StepMania-level creators and players. So far, only level generation \citep{donahue2017dance, tsujino2018dance}, and difficulty prediction \citep{tsujino2019characteristics, caronongan2021predicting} have been applied to this data. However, other tasks or subtasks of the former utilizing this data may also be interesting. Examples include detecting salient events in music, anomaly detection, and early anomaly detection. StepMania data is, in essence, extensively labeled sound data, and we encourage using it as such.

\paragraph{Our contributions.} We propose using OR to predict the difficulties of StepMania levels. Our contributions include the following:
\begin{itemize}
    \item We provide the first analysis of deep OR methods on the task of difficulty prediction on StepMania data, resulting in a new state of the art for this task. (Sec. \ref{sec:ORexperiments})
    \item We increase the number of standardized data sets and expand upon previous data sets, provide a fundamental description of the data, and extensive data analysis about StepMania data and the relationships between different StepMania data sets. (Sec. \ref{sec:datasets}, \ref{sec:ORcrossdatasets}, \ref{sec:rankingexperiments})
    \item Finally, we demonstrate that OR models can improve human labels. For this, we evaluate each model considered here on its accuracy on user rankings of pairs of StepMania levels. (Sec. \ref{sec:user-test})
\end{itemize}

\section{Related Work}
We cover related work starting with previous work on StepMania and closely related data. Then, OR surveys and OR taxonomies are discussed from an ML and social sciences perspective.

\subsection{StepMania and Related Data}
\citet{donahue2017dance} first investigated ML on StepMania data.
They used ML for the task of level generation or learning to choreograph.
\citet{tsujino2018dance} improved upon this previous work by blending more challenging and less challenging levels, creating levels of intermediary difficulty.
\citet{halina2021taikonation} transferred this general approach to another rhythm game, Taiko no Tatsujin.

\citet{tsujino2019characteristics} first investigated the task of level difficulty analysis or difficulty estimation of StepMania levels. They clustered levels based on automatically extracted hand-picked features and found the resulting clusters to correlate with difficulty levels.
\citet{caronongan2021predicting} built on this idea by combining features calculated for a level by StepMania with the count and required speed of specific patterns occurring in the levels to predict the level difficulty using a classification approach.

Broadening the perspective from just StepMania levels to musical charts, which are very similar data, \citet{sebastien2012score} distinguished four difficulty levels for piano pieces, from beginner to virtuoso. Musical features, including playing speed, chord ratio, and fingering difficulty, were automatically extracted and then separately classified and aggregated via human-validated decision rules.
Similarly, \citet{chiu2012study} also extracted features from symbolic music charts but used various regression methods to relate these features to nine different difficulty levels. Finally, \citet{ghatas2022hybrid} developed a hybrid approach by combining the top-5 features from \citet{chiu2012study} with a deep convolutional NN based on piano roll representation of the music.

In contrast to existing work, we apply methods from OR to the difficulty estimation of StepMania levels. For this reason, we also list related work on OR.

\subsection{Ordinal Regression Surveys}
\citet{gutierrez2015ordinal} conducted a comprehensive survey and experimental study of various OR models, which we use as a reference to choose methods to compare. \citet{tutz2022ordinal} recently provided a taxonomy of OR models and applied these to a case study about public fear of nuclear energy. Similarly, \citet{burkner2019ordinal} describe OR in detail from a social sciences perspective.

Contemporary research shows that OR methods generally outperform their non-OR counterparts on OR problems.
\citet{niu2016ordinal} compare previous shallow and deep non-OR models with shallow and deep OR models and demonstrate the superiority of OR on age estimation data and the need for deep models in this field. \citet{fu2018deep} compare previous deep convolutional non-OR approaches against a deep OR approach, again demonstrating deep OR superiority. \citet{saad2019twitter} provide the first application of OR combined with four ML models to a sentiment analysis task.

\section{StepMania Difficulty as Ordinal Regression}
\label{sec:background}

We first describe the problem setting. Then, we argue conceptually why OR is the superior approach for our problem. Lastly, we describe a selection of OR methods that we consider most relevant to the analysis performed herein.

We want to train an ML model that estimates the difficulty of a StepMania level, which is our data sequence $x_i=(x_i^{(1)},\dots,x_i^{(d)})$ of length $d$. We specify the $x_i^{(j)}$ in the experiments (Sec. \ref{sec:experiments}). StepMania level difficulties are natural numbers and our labels $y_i \in \{1, \dots, K\}$. We also want prediction errors to be as small as possible. Let $\hat{y}$ be a model's prediction. We assign a cost $c(\hat{y},y):=|\hat{y}-y|$ to a prediction with the correct label $y$. Training a model that minimizes this cost is an OR problem.

Readers might wonder why OR is separate from standard classification or regression. Later in the experiments, we will consider these as non-OR baselines. However, it is notable that both classification and regression do not inherently fit the metric present in OR, i.e., the cost function $c$. Usual classification approaches will minimize the negative log-likelihood (NLL). Based on this, the classifier loss function is independent of the ordering of labels, meaning an off-by-one error is treated the same as an entirely wrong prediction. Regression might seem more suitable than classification. However, the whole-valued nature of our labels complicates this approach. Training a regular regressor minimizing mean absolute error (MAE) without rounding to the nearest integer will result in the regressor not taking the rounding threshold into account.

In the following, we describe the OR methods considered in this work. We chose them based on previous studies \citep{gutierrez2015ordinal} and their compatibility with neural network architectures.

\paragraph{NNRank.} \citet{cheng2008neural} proposed using a set of binary classifiers to solve OR problems. Specifically, a model predicts $K-1$ binary classifiers, where the $i$-th binary classifier predicts the probability that $i < y$. In this sense, the target of an input is a vector of $y-1$ ones followed by $K-y$ zeros, i.e., for $y=4$, the target is $t=(1,1,1,0,\dots, 0)$.

\paragraph{RED-SVM.} \citet{lin2012reduction} introduced the reduction-based support vector machine (SVM) for OR. They propose to reduce multiple binary classifiers (as in NNRank) to just one binary classifier. For an input $x$, target $y$, and a category $k$, this classifier decides whether $k < y$.
In practice, one data point $(x,y)$ is transformed into $K-1$ data points $((x,1),1<y), ((x,2),2<y), \dots, ((x,K-1),K-1<y)$. An SVM is then trained on this data. In the original SVM formulation, this translates to learning a linear (or kernelized) regressor and $K-1$ thresholds.

\paragraph{Laplace.} \citet{diaz2019soft} proposed using soft labels. Instead of learning to predict a classification target vector $t=(0,\dots, 0, 1, 0, \dots, 0)$, they propose learning a smoothed target vector using a distance metric $\phi(y, i)$,
\begin{equation}
t_i=\frac{\exp^{-\phi(y,i)}}{\sum_{k=1}^K \exp^{-\phi(y,k)}}.
\end{equation}
These targets are discretely sampled common probability distributions normalized by the denominator. For $\phi(y, i)=|y-i|$, the distribution is a Laplace distribution, and for $\phi(y, i)=\lVert y-i\rVert^2$, the distribution is normal. The cross-entropy between target and softmax predictions of a model is then used as training loss.

\paragraph{Binomial.} With soft labels as distributions in mind, we also propose using a Poisson binomial distribution with $K+3$ Bernoulli trials, mean $(K+3)p=y+1$, and variance $1$ as a target. More specifically, we choose $y+1$ Bernoulli trials with probability $p_1$ and $K-y+2$ Bernoulli trials with probability $p_2$, where
\begin{align}
p_1&:=\frac{y+1}{K+3}+\sqrt{\frac{(K-y+2)^2}{(K+3)^2}-\frac{(K-y+2)}{(y+1)(K+3)}}\\
p_2&:=\frac{y+1}{K+3}-\sqrt{\frac{(y+1)^2}{(K+3)^2}-\frac{(y+1)}{(K-y+2)(K+3)}}.
\end{align}
These choices of $p_1$ and $p_2$ result in a mean of $y+1$ and a variance of $1$ if the variance of a binomial with $p=\frac{y+1}{K+3}$ is larger or equal to one. This is guaranteed by considering $K+3$ Bernoulli trials instead of $K-1$. We discuss this in greater detail in the supplementary material.

\section{StepMania Level Data}
\label{sec:datasets}
StepMania\footnote{\url{https://www.stepmania.com/}} is an open-source game engine with over 100 contributors initially developed as a clone of Konami's arcade game series Dance Dance Revolution (DDR). StepMania has become the engine for multiple games based on DDR, including In the Groove, Pump It Up Pro, and others. In StepMania, a player may choose a song and a difficulty level to ``play''. This starts a game level in which the player must step onto four arrows, left, right, up, and down, to a certain rhythm on a controller on the floor, which is typically called a pad. Usually, this rhythm is in line with the chosen song playing in the background. The act of playing resembles dancing, where DDR likely got its name. Levels appear in combination with songs and it is most common for authors to distribute songs in packs, which is a collection of multiple songs. Most packs are created by a single individual or a small group working closely together, guaranteeing a homogeneous difficulty interpretation of levels in a pack.

Each song is associated with a music file and a background image or video. An SM (for StepMania) file encodes the level data in a custom ASCII-based file format. SM is an elementary file format used in StepMania and other rhythm games. A newer file format SSC with more design features has been established from StepMania version 5 (previously known as StepMania Spinal Shark Collective fork) and onward. Even though SSC contains more options for level design, more songs are available in the SM format (newer levels usually encode both SM and SSC), so we will use this format for this work.
An SM file starts with header information followed by encoding at least one level. The header contains features that are consistent across levels, such as the title, artist, and tempo changes. A sequence then describes a level by dividing the song into measures (from musical notation), encoding which inputs are required at what time. Each measure can be split into 4--192 equidistant parts that define the granularity in which notes can be assigned. Together with the tempo, this defines the maximum speed at which inputs may need to be entered and the possible rhythmic complexity.

From an ML perspective, there are a few complications with this data. Packs are created for personal enjoyment rather than ML purposes, leading to categories with very few or no samples. Additionally, multiple levels per song with different labels may be available. These different levels for the same song are correlated both naturally and intentionally by the design process of the authors.
This needs to be considered when splitting data into training and test sets.
These issues are addressed in the experiments (Sec.~\ref{sec:experiments}).

Packs of songs for StepMania are freely available online, facilitating easy access to labeled data. StepMania Online\footnote{\url{https://search.stepmaniaonline.net/}} is one of these repositories, searchable by pack name but also song names or authors. Crawling the packs available there results in 602 GB of data, although this also includes the music file and, in some cases, background videos, which we discard in our analysis. We also consider packs that are released in other communities. Zenius-I-vanisher\footnote{\url{https://zenius-i-vanisher.com}}, for instance, has a relatively active community. Another notable community that was very influential in the proliferation, organization of conventions, and tournament-play of StepMania is DDRIllini\footnote{\url{https://ddrillini.club/}}.

Some prolific individuals or groups create multiple packs, which we collate into more extensive data sets with consistent themes. This is how the following data sets were chosen. Tab.~\ref{tab:datasets} provides an overview of all data sets.
\paragraph{ITG.} \citet{donahue2017dance} introduced the In The Groove (ITG) data set made up of 133 songs or 652 levels from multiple authors, which combine the packs ITG1 and ITG2. We expand this ITG data set by adding ITG3 and ITGRebirth, yielding 297 songs and 1,469 levels. ITG primarily contains electronic indie music. In contrast to the other data sets, ITG originates from the game studio Roxor Games and has been slightly modified by the StepMania community over time. Thus ITG is not purely a community contribution.

\paragraph{fraxtil.} \citet{donahue2017dance} also introduced the fraxtil data set, which combines three packs from one level author, known as fraxtil, and contains 90 songs and 450 levels. fraxtil primarily contains electronic music.

\paragraph{Gpop.} We propose Gpop, encompassing 542 songs and 2,710 levels from twelve packs. The levels are almost exclusively created by creator Gpop. Gpop features mostly Japanese pop and Vocaloid music.

\paragraph{Gulls.} We propose Gull's Arrows, which consists of seven packs by creator Gamergull with ten songs each, for a total of 70 songs and 260 levels. The songs in this data set include mostly electronic music and some video game soundtracks.

\paragraph{Speirmix.} Lastly, we propose using Ben Speirs' Speirmix Galaxy, consisting of 267 songs and 1,185 levels written chiefly by the creator Ben Speir. The genre of Speirmix mainly focuses on modern music, featuring many pop songs from the 2000s and 2010s.

\begin{table}[tb]
    \centering
    
    \begin{tabular*}{\linewidth}{@{\extracolsep{\fill}}lrrrr}
        \toprule
        Name & \#Packs & \#Songs & \#Levels & Min-Max \\
        \midrule
        $\text{ITG}^1$ & 4 & 297 & 1,469 & 1-14\\
        fraxtil & 3 & 90 & 450 & 1-15\\
        $\text{Gpop}^2$ & 12 & 542 & 2,710 & 1-18\\
        $\text{Gulls}^2$ & 7 & 70 & 260 & 2-14\\
        $\text{Speirmix}^2$ & 1 & 267 & 1,185 & 1-15\\
        \bottomrule
    \end{tabular*}
    \caption{Overview of all data sets with statistics. Statistics shown are the number of packs used for construction, the number of songs included, the number of levels included, and the minimum and maximum difficulty available in each data set. ${}^1$ indicates expanded previous data sets. ${}^2$ indicates newly proposed data sets.}
    \label{tab:datasets}
\end{table}

The distinctions in StepMania-play are more complicated than they have appeared so far. The StepMania community is split into multiple, sometimes overlapping, groups. The differences between these groups vary from the type of controller used to the design style of levels. Examples include (1) ITG, a common synonym for StepMania.
(2) DDR, which is based on the original DDR style levels. (3) Keyboard play, which uses a computer keyboard as a controller. (4) Pump it Up, which uses a controller with up-right, down-right, down-left, up-left, and center keys. (5) Rhythm Horizon, a newer development, which uses all directions mentioned so far.
Additionally, these communities have been split into different genres at different times. Currently, there are three such sub-genres for StepMania play. These groups referred to themselves as (1) Tech, which includes players focussing on accuracy in their play. (2) Stamina, which includes players that focus on long levels with continuous streams, i.e., continuous single steps with no breaks. (3) Modding, which includes players that play levels where reading is the main difficulty, meaning that each level's visuals are modified to such an extent that the required arrows are significantly harder to detect on screen. This work focuses primarily on pad-play (as opposed to keyboard play) and tech levels. The data sets also primarily feature older content starting around 2005. We expect the methodology proposed here to also function on data from most other groups. However, Modding levels would need entirely different approaches, most likely including extracting video to determine the difficulty of levels.

\section{Experiments}
\label{sec:experiments}

\paragraph{Data sets.}
As mentioned previously, the data sets are unbalanced, with some difficulty levels being rare. We first deal with small categories by joining adjacent difficulty categories until every remaining category accounts for at least 2 percent of all levels. We balance the data sets as part of the training procedure and evaluation protocol described later.

Secondly, the data sets described here must be split into training and test sets. We do not use regular cross-validation since multiple levels of the same song are correlated. Instead, we use Monte Carlo cross-validation with rejection sampling. Specifically, we choose 20\% of songs (with all their respective levels) as a test set and the remainder as the train set. We reject samples until all difficulties are present in both sets. With this process done, we have an approximate 80-20 train-test split with each label present in both sets. We repeat this Monte Carlo cross-validation process 100 times yielding 100 different train-test splits for each data set.

\paragraph{Feature extraction.}
From the SM files, we extract each level sequence separately. We encode each sequence element as a 19-dimensional feature vector $x_i^{(j)}\in \mathbb R^{19}$. This vector contains the tempo (1 feature), an encoding of the note level (7 features), level progress in time (1 feature), level progress in sequence length (1 feature), time since the last element (1 feature), a one-hot encoding of the step direction in case of a tap (4 features) and hold (4 features), for a total of 19 features. Due to their rarity, we ignore all other potential features of the level sequence, like tempo changes, mines, or other effects. Find a more detailed description in the supplement.

\paragraph{Compared Methods.} As our baselines, we evaluate PATTERN proposed by \citet{caronongan2021predicting} as a non-deep non-OR method and a classification and regression model as deep non-OR methods. We compare all methods from Sec. \ref{sec:background} to these baselines.

\paragraph{Model architecture.}
For PATTERN, we use a three-layer multilayer perceptron with a hidden size of 32. This model only uses a feature vector instead of a time series.
Building on the recent success in sequence processing with transformers \citep{vaswani2017attention, NEURIPS2020_1457c0d6}, we use the encoder of a transformer as the backbone to all deep models. Specifically, we use three layers of dimensionality 64 with four attention heads. Before applying the encoder, a small convolutional layer of kernel size two projects the $19$ input features to the embedding dimension. Then, we add a positional encoding. Global average pooling reduces the time series to a single vector before each model's head. We ensemble eight random sub-samples of length 60 from each input sequence to produce a prediction.

A model's head depends on the chosen method. We train PATTERN and the classifier with NLL. The regressor is trained using MAE. NNRank consists of $K-1$ logits trained with NLL. We evaluate each data point for every threshold in RED-SVM and train the binary classifier with NLL. The target is a discrete distribution for Laplace, so we use cross-entropy between the target and a softmax multi-class head. For Binomial, we do the same as for Laplace, except that we replace the Laplace target distribution with a Poisson binomial distribution. Find more details in the supplement.

\paragraph{Training.}
We train each model on the training set for 200 epochs with a batch size of 128, AdamW \citep{loshchilov2018decoupled} as the optimizer with a weight decay of 5e-2, and a learning rate of $lr = \frac{\# levels}{1500}\cdot 10^{-4}$ to adapt for data set size.
Due to the unbalanced data sets, we use a weighted random sampler to select data points, resulting in each difficulty being drawn with equal probability.
We train one model for each method on each training set of all 100 train-test splits of each data set. This results in a total of 3,500 models.

\subsection{Evaluating OR Methods on StepMania-Level Difficulty Estimation}
\label{sec:ORexperiments}
\paragraph{Metrics}
Due to the unbalanced data sets, we consider a class-weight-normalized version of the MAE as our metric. We refer to our metric as the weighted absolute error (WAE). The WAE of a model $f$ on a data set $D$ with data points $(x,y)$ and $K$ classes is defined as
\begin{align}
    \text{WAE}_D(f) = \frac{1}{K} \sum_{(x,y) \in D} \frac{1}{w_y} |y-f(x)|\\
    \text{with } w_y = |\{\mathbf{x} | (\mathbf{x},y) \in D\}|.
\end{align}
Notably, for a balanced data set, WAE and MAE are equal.

We evaluate each of the 3,500 trained models using WAE and compute the mean and standard deviation across the 100 Monte Carlo cross-validation samples. The results can be found in Tab. \ref{tab:WAEeval}a. The supplementary material contains additional evaluations for other metrics, including MAE.

\begin{table*}[tb]
    \begin{subtable}[h]{\textwidth}
    \centering
    \begin{tabular*}{\textwidth}{@{\extracolsep{\fill}}llccccc} 
 \toprule
    && ITG & fraxtil & Gpop & Gulls & Speirmix\\ 
    \midrule
    \multirow{3}{*}{non-OR}&PATTERN         & 0.457 $\pm$ 0.043 & 0.736 $\pm$ 0.092 & 0.460 $\pm$ 0.028 & 0.695 $\pm$ 0.101 & 0.297 $\pm$ 0.037\\ 
    \cmidrule{2-7}
    &Classification  & \textbf{0.366 $\pm$ 0.033} & 0.480 $\pm$ 0.080 & \textbf{0.342 $\pm$ 0.027} & \underline{0.273 $\pm$ 0.063} & \underline{0.274 $\pm$ 0.046}\\ 
    &Regression      & 0.379 $\pm$ 0.030 & 0.489 $\pm$ 0.058 & 0.356 $\pm$ 0.025 & 0.284 $\pm$ 0.064 & 0.278 $\pm$ 0.043\\ 
    \midrule
    \multirow{4}{*}{OR}&NNRank          & \underline{0.372 $\pm$ 0.033} & \textbf{0.444 $\pm$ 0.065} & \underline{0.344 $\pm$ 0.025} & \underline{0.275 $\pm$ 0.066} & \underline{0.268 $\pm$ 0.047}\\ 
    &RED-SVM         & \underline{0.368 $\pm$ 0.030} & 0.481 $\pm$ 0.058 & \underline{0.349 $\pm$ 0.024} & \textbf{0.262 $\pm$ 0.057} & \underline{0.268 $\pm$ 0.043}\\ 
    &Laplace         & \underline{0.367 $\pm$ 0.029} & \underline{0.455 $\pm$ 0.070} & \underline{0.342 $\pm$ 0.026} & \underline{0.270 $\pm$ 0.067} & \underline{0.269 $\pm$ 0.051}\\ 
    &Binomial        & \underline{0.368 $\pm$ 0.031} & \underline{0.448 $\pm$ 0.072} & \underline{0.344 $\pm$ 0.027} & \underline{0.264 $\pm$ 0.069} & \textbf{0.265 $\pm$ 0.047}\\ 
\bottomrule \end{tabular*}
    \caption{Trained and evaluated on the same data set.}
    \end{subtable}
    \begin{subtable}[h]{\textwidth}
    \centering
    \begin{tabular*}{\textwidth}{@{\extracolsep{\fill}}llccccc} 
    \toprule
    && ITG & fraxtil & Gpop & Gulls & Speirmix\\
    \midrule
    \multirow{3}{*}{non-OR}&PATTERN         & 0.717 $\pm$ 0.16& 0.618 $\pm$ 0.179& 0.717 $\pm$ 0.167& 0.549 $\pm$ 0.126& 0.670 $\pm$ 0.129\\
    \cmidrule{2-7}
    &Classification  & \underline{0.461 $\pm$ 0.060}& 0.477 $\pm$ 0.058& 0.488 $\pm$ 0.060& \textbf{0.379 $\pm$ 0.085}& 0.510 $\pm$ 0.060\\
    &Regression      & 0.483 $\pm$ 0.048& \underline{0.469 $\pm$ 0.078}& \underline{0.480 $\pm$ 0.080}& 0.411 $\pm$ 0.122& 0.562 $\pm$ 0.103\\
    \midrule
    \multirow{4}{*}{OR}&NNRank          & \textbf{0.455 $\pm$ 0.046}& \underline{0.461 $\pm$ 0.065}& \underline{0.469 $\pm$ 0.064}& \underline{0.403 $\pm$ 0.107}& \textbf{0.472 $\pm$ 0.089}\\
    &RED-SVM         & \underline{0.465 $\pm$ 0.046}& \textbf{0.452 $\pm$ 0.074}& \textbf{0.465 $\pm$ 0.072}& \underline{0.387 $\pm$ 0.104}& 0.546 $\pm$ 0.095\\
    &Laplace         & 0.475 $\pm$ 0.057& \underline{0.458 $\pm$ 0.074}& \underline{0.469 $\pm$ 0.069}& \underline{0.388 $\pm$ 0.108}& 0.529 $\pm$ 0.095\\
    &Binomial        & 0.475 $\pm$ 0.057& \underline{0.457 $\pm$ 0.071}& \underline{0.471 $\pm$ 0.066}& \underline{0.395 $\pm$ 0.103}& 0.498 $\pm$ 0.080\\
    \bottomrule
    \end{tabular*}
    \caption{Trained on separate data sets from the one being evaluated.}
    \end{subtable}
    \caption{OR and classification models outperform feature-extraction-based PATTERN and regression models. This table shows WAE performance rounded to the third nearest digit for different models on all StepMania data sets averaged over 100 models trained on separate Monte Carlo cross-validation splits. Lower values are better. The smallest value in each column is \textbf{bold}. According to a standard 5\% significance t-test, all values in a column insignificantly higher than the best value are \underline{underlined}.}
    \label{tab:WAEeval}
\end{table*}

\begin{figure}[!tb]
    \includegraphics[width=\linewidth]{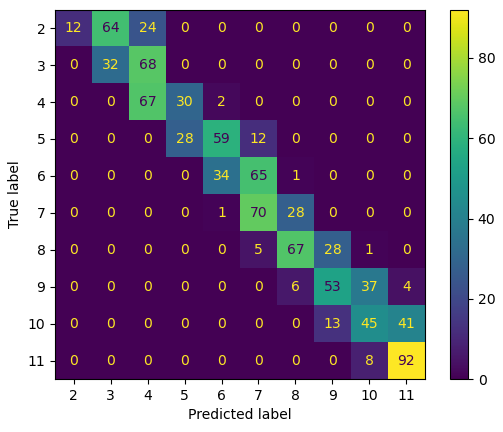}
    \caption{Difficulty depends on subjective perceptions of level creators. This confusion matrix shows that there is often an offset of up to one category if a model trained on one dataset (here Speirmix) is evaluated on another dataset (here Gulls). The confusion matrix is averaged and category-normalized. The model was trained with the Binomial loss.}
    \label{fig:confusion}
\end{figure}

\subsection{Difficulty Across Data Sets}
\label{sec:ORcrossdatasets}
Readers might wonder whether the different data set's labels are consistent when compared to one another. They are not consistent. See the confusion matrix in Fig. \ref{fig:confusion}, where Binomial was trained on Speirmix but evaluated on Gulls, for a visualization that difficulties can be offset significantly. More confusion matrices can be found in the supplementary material. With this in mind, we evaluated for each data set pair $(A, B)$ how a model trained on $A$ performs on $B$.
Difficulties from $B$ are adjusted based on the pooling defined for $A$.

Specifically, for each pair of data sets $(A, B)$ and method $M$, all 100 models trained on $A$ for method $M$ are evaluated on the entire data set $B$ and averaged. This would produce a $5\times 5\times 7$ tensor of evaluations. To make these evaluations easier to compare to Tab. \ref{tab:WAEeval}a, we present for a data set $B$ and method $M$ the mean and standard deviation of method M trained on all data sets but $B$. Tab. \ref{tab:WAEeval}b contains this evaluation. Find the entire tensor of evaluations in the supplementary material.

\subsection{Difficulty Prediction as a Ranking Problem}
\label{sec:rankingexperiments}

As shown in the previous section, the previous evaluations are structurally biased because the given difficulty levels are inconsistent between data sets. However, viewing the data sets as ranking problems will avoid this bias. Specifically, instead of evaluating whether a data point $x$ is predicted as label $y$, we evaluate for a pair of data points $x,x'$ with $y<y'$ whether the prediction of $x'$ is larger than $x$. Moving from OR models to ranking models is simple, as OR predictions can be compared to get ranking predictions. From a data set $D$ with data points $(x,y)$ we construct a new data set $D' := \{((x,x'),sign(y-y'))|(x,y), (x',y') \in D)\}$. The sign function can then be considered $+1$, $-1$, or $0$ for equal labels affecting data set construction. We present the results here, assuming a separate label $0$ for equal labels.

On this data set $D'$, we can then measure accuracy. A slight inaccuracy might occur if either the data set or the predictions claim that a pair of data points is equal since the difficulties still do not align. For this reason, we will only consider the accuracy for pairs that agree or disagree without equality. To avoid confusion with previous experiments, we will refer to this metric as the agreement of the ranking predictions with the ranked labels or just agreement. We repeat both previous experiments using agreement. Models trained on a train set are evaluated on their respective test set (all pairs are drawn just from the test set), and models trained on all other data sets are evaluated on the one data set can be found in Tab. \ref{tab:agreement}.

\begin{table*}[tb]
    \begin{subtable}[h]{\textwidth}
    \centering
    \begin{tabular*}{\textwidth}{@{\extracolsep{\fill}}llccccc} 
 \toprule
&& ITG& fraxtil& Gpop& Gulls& Speirmix\\
\midrule
\multirow{3}{*}{non-OR}&PATTERN & 0.988 $\pm$ 0.003 & 0.981 $\pm$ 0.005 & 0.990 $\pm$ 0.001 & 0.971 $\pm$ 0.009 & 0.995 $\pm$ 0.001\\ 
\cmidrule{2-7}
 &Classification & \textbf{0.992 $\pm$ 0.002} & 0.991 $\pm$ 0.003 & \underline{0.994 $\pm$ 0.002} & 0.996 $\pm$ 0.003 & 0.995 $\pm$ 0.002\\ 
 &Regression & \underline{0.992 $\pm$ 0.002} & 0.992 $\pm$ 0.003 & \underline{0.994 $\pm$ 0.001} & \underline{0.996 $\pm$ 0.003} & \underline{0.996 $\pm$ 0.002}\\ 
 \midrule
 \multirow{4}{*}{OR}&NNRank & \underline{0.992 $\pm$ 0.002} & \textbf{0.992 $\pm$ 0.003} & \textbf{0.994 $\pm$ 0.001} & \textbf{0.997 $\pm$ 0.002} & \textbf{0.996 $\pm$ 0.002}\\ 
 &RED-SVM & \underline{0.992 $\pm$ 0.002} & 0.992 $\pm$ 0.003 & \underline{0.994 $\pm$ 0.001} & \underline{0.997 $\pm$ 0.003} & \underline{0.995 $\pm$ 0.002}\\ 
 &Laplace & \underline{0.992 $\pm$ 0.002} & \underline{0.992 $\pm$ 0.003} & \underline{0.994 $\pm$ 0.001} & 0.996 $\pm$ 0.003 & \underline{0.996 $\pm$ 0.002}\\ 
 &Binomial & \underline{0.992 $\pm$ 0.002} & \underline{0.992 $\pm$ 0.003} & \underline{0.994 $\pm$ 0.002} & 0.996 $\pm$ 0.003 & \underline{0.996 $\pm$ 0.002}\\ 
\bottomrule \end{tabular*}
    \caption{Trained and evaluated on the same data set.}
    \end{subtable}
    \begin{subtable}[h]{\textwidth}
    \centering
    \begin{tabular*}{\textwidth}{@{\extracolsep{\fill}}llccccc} 
 \toprule
&& ITG& fraxtil& Gpop& Gulls& Speirmix\\ 
\midrule
\multirow{3}{*}{non-OR}&PATTERN& 0.980 $\pm$ 0.006& \underline{0.985 $\pm$ 0.009}& 0.978 $\pm$ 0.008& \underline{0.990 $\pm$ 0.005}& 0.992 $\pm$ 0.005\\
\cmidrule{2-7}
&Classification& \underline{0.988 $\pm$ 0.004}& 0.991 $\pm$ 0.002& \underline{0.990 $\pm$ 0.003}& \underline{0.995 $\pm$ 0.002}& 0.995 $\pm$ 0.000\\
&Regression& \underline{0.989 $\pm$ 0.002}& \underline{0.992 $\pm$ 0.001}& \underline{0.992 $\pm$ 0.001}& \underline{0.995$\pm$ 0.001}& \underline{0.996 $\pm$ 0.000}\\
\midrule
\multirow{4}{*}{OR}&NNRank& \textbf{0.990 $\pm$ 0.002}& \textbf{0.993 $\pm$ 0.001}& \textbf{0.992 $\pm$ 0.001}& 0.995 $\pm$ 0.002& \textbf{0.996 $\pm$ 0.000}\\
&RED-SVM& \underline{0.989 $\pm$ 0.002}& \underline{0.993 $\pm$ 0.001}& \underline{0.992 $\pm$ 0.001}& \textbf{0.995 $\pm$ 0.001}& 0.996 $\pm$ 0.000\\
&Laplace& \underline{0.989 $\pm$ 0.004}& \underline{0.993 $\pm$ 0.001}& \underline{0.991 $\pm$ 0.002}& \underline{0.995 $\pm$ 0.002}& \underline{0.996 $\pm$ 0.000}\\
&Binomial& \underline{0.988 $\pm$ 0.004}& \underline{0.993 $\pm$ 0.001}& \underline{0.991 $\pm$ 0.002}& \underline{0.995 $\pm$ 0.002}& \underline{0.996 $\pm$ 0.000}\\
\bottomrule \end{tabular*}
    \caption{Trained on separate data sets from the one being evaluated.}
    \end{subtable}
    \caption{Generalization is near perfect when considering StepMania difficulty prediction as a ranking problem. This table shows agreement of different models on all StepMania data sets averaged over 100 models trained on separate Monte Carlo cross-validation splits. Higher values are better. The largest value in each column is \textbf{bold}. According to a standard 5\% significance t-test, all values in a column insignificantly lower than the best value are \underline{underlined}.}
    \label{tab:agreement}
\end{table*}

\subsection{Does ML Improve the Original Difficulties?}
\label{sec:user-test}

We expect the labels created by authors for StepMania levels to be noisy. With this motivation in mind, we evaluate whether the predictions of the models considered here improve upon the labels from the original authors using user feedback. Since StepMania players need context to rate levels, we decided to use the previously mentioned ranking approach to evaluate agreement.

\paragraph{Experimental Setup.} We provide a participant with two songs and ask them which is more difficult. In essence, we ask the players to label a pair of StepMania levels $(a,b)$ with a binary label ($\text{rank}(a)>\text{rank}(b)$?).

We are left with a choice of which pairs to evaluate. Given the large number of possible pairs and the expected low number of user evaluations, we choose the pairs carefully. Specifically, we choose pairs for which different models or the original labels disagree.

Similarly to the previous experiment, while our model predictions might indicate that two StepMania levels have the same difficulty, players will always consider one level easier. As with the previous experiment, we disregard equal predictions entirely from the evaluation, i.e., the data set only includes pairs for which the predictions are not equal since players will always disagree with equality.

Conducting this experiment, we collected 217 human labels containing 105 unique pairs. Most pairs were evaluated more than once, and different players sometimes disagreed on what order they should have. We assigned each ordered pair a correctness value between 0 and 1, corresponding to the average support for this ordering. This correctness is reflected in the agreement. Tab. \ref{tab:user-test} shows models evaluated on a separate data set from the one they were trained on. The results evaluating a model trained and evaluated on the same data set can be found in the supplement.

\begin{table}[tb]
    \centering
    \resizebox{\linewidth}{!}{
     \begin{tabular}{lrrrrr} 
     \toprule
    & ITG& fraxtil & Gpop & Gulls & Speirmix\\
    \midrule
    Original&  \textbf{0.675} & 0.470 & 0.257 & 0.607 & 0.388\\\midrule
    PATTERN& 0.507& 0.545 & 0.499 & 0.309 & 0.649\\
    \midrule
    Classification& 0.322 & 0.659 & \textbf{0.726} & \textbf{0.721} & 0.746\\
    Regression& 0.435 & 0.675 & 0.718 & 0.682 & 0.736\\
    \midrule
    NNRank& 0.384 & 0.664 & 0.716& 0.701& 0.734\\
    RED-SVM& 0.446 & \textbf{0.678} & 0.712 & 0.680 & 0.744\\
    Laplace& 0.418 & 0.669 & 0.713 & 0.702 & \textbf{0.747}\\
    Binomial& 0.409 & 0.667 & 0.711& 0.692 & 0.743\\
    \bottomrule \end{tabular}}
    \caption{Most models improve upon the original labels of each data set, except for ITG. This table shows the agreement with user evaluations on various data sets of models trained on other data sets.}
    \label{tab:user-test}
\end{table}

\section{Discussion \& Conclusion}
According to Tab. \ref{tab:WAEeval}a, all OR methods, as well as just a basic classification approach, perform very well. However, no one method can be determined to be the best. In contrast to the varied results among the OR methods and classification, a basic regression approach and the PATTERN method from \citet{caronongan2021predicting} perform poorly. The PATTERN method performs subpar in all experiments, likely because it is limited to only a couple of static features extracted from the level sequences. Regression performs close to the better methods, even though it is significantly outperformed in most experiments. We do not know why classification generally performs well, whereas regression does not.

Viewing StepMania data as a ranking task instead of an OR task demonstrates a near-perfect generalization as shown in Tab. \ref{tab:agreement}.
Comparing Tab. \ref{tab:WAEeval}a to Tab. \ref{tab:WAEeval}b, however, highlights that these methods do not generalize from one data set to another when viewed naively. From this, we can follow that the poorer performance in Tab. \ref{tab:WAEeval}b is due to labeling inconsistencies among authors of these packs. Fig. \ref{fig:confusion} exemplifies this by demonstrating an offset of up to one between the difficulty scales of the Speirmix and Gulls data sets.

Finally, considering Tab. \ref{tab:user-test}, we demonstrate that all OR models can unify different StepMania packs into one difficulty ranking. Except for ITG, all methods improve upon the original authors labeling. The better performance of the original labels for ITG is likely due to the ITG packs originating from a professional game studio instead of private individuals. We found that most errors made as part of Tab. \ref{tab:user-test} are because the model is oblivious to some StepMania-level features. The models investigated herein are limited by the feature extraction performed on the SM files. Some rare design elements were entirely ignored. Specifically, we ignored (1) mines, which require the player to step off of a direction not to trigger it, (2) tempo changes, which change the tempo during play, (3) warps, which skip ahead in a level, (4) stops, which stop the progress of the song for a predetermined time, and many more. These effects do affect the difficulty of StepMania levels. However, due to their rarity, they can lead to overfitting during training. Additionally, higher difficulties might also be caused by salient events in the music, which were not considered herein, being hard to detect, or by a level's steps being aligned not with the lead beat but with some salient events in the background.

Using all available StepMania data simultaneously during training is a significant challenge due to the labels not aligning, which is left as future work. The OR models considered here are not created equal, considering their practical use. RED-SVM specifically learns a regressor, which can be used to rank even where other methods consider a pair equal. This, in addition to its generally good performance, is why we suggest using RED-SVM to construct a unifying difficulty estimator for StepMania. Assuming RED-SVM is trained in a multi-task fashion on a large set of StepMania packs, each sharing a regressor but with separate thresholds, the final regressor can produce difficulty estimates on any of the trained scales. Additionally, scales can be hand-designed using standard StepMania levels as a reference. Considering that most levels are for mid-level play, a finer scale could split these mid-difficulty levels.

Based on our experiments, we conclude that StepMania-level difficulties are noisy and that ML models can help remove this noise and define a standard difficulty scale. Given its complexity, we also conclude that StepMania data should provide an excellent future test bed for deep OR models.

\clearpage
\appendix
\section*{Acknowledgments}
The authors acknowledge support by the Carl-Zeiss Foundation, the DFG awards KL 2698/2-1, KL 2698/5-1, KL 2698/6-1, KL 2698/7-1, BU 4042/2-1, and BU 4042/1-1, the BMBF awards 03|B0770E, 01|S21010C, and 01|S20048, and the BMWK award KEEN (01MK20014U,01MK20014L). The authors would also like to thank all the contributors of the user experiment.

\bibliographystyle{named}
\bibliography{refs}

\clearpage

\section{Ordinal Regression Methods}

\paragraph{NNRank.} A model predicts $K-1$ binary classifiers \citep{cheng2008neural}. The $i$-th binary classifier predicts the probability that $Pr[i < y]$. The target of an input is a vector of $y-1$ ones followed by $K-y$ zeros, i.e., for $y=4$, the target is $t=(1,1,1,0,\dots, 0)$. An element of this target $t_i$ is the target for the $i-th$ binary classifier. The binary classifiers can be trained with any method. We choose to use binary cross entropy as our loss. A prediction from these binary classifiers will take the form of a vector of probabilities $p=(p_1,\dots,p_{K-1})$. There are different approaches to predicting a final category. We choose the largest $i$ with probability larger than $0.5$:
\begin{equation}
\begin{array}{ll}
\displaystyle\max_y &y,\\
\text{s.t.}& p_y\geq 0.5
\end{array}
\end{equation}

\paragraph{RED-SVM.}  For an input $x$ and a category $k$ we learn a model that decides whether $k < y$ \citep{lin2012reduction}. One data point $(x,y)$ is transformed into $K-1$ data points $((x,1),1<y), ((x,2),2<y), \dots, ((x,K-1),K-1<y)$.
Denote the data points as $x_1,\dots,x_n$, their original labels as $y_1,\dots,y_n$ and their new labels as $y_1^{(1)}, \dots, y_1^{(K-1)},\dots,y_n^{(1)}, \dots, y_n^{(K-1)}$, where $y_i^{(j)}:=2[[j<y_i]]-1$. Note, that each data point $x_i$ is associated with multiple new labels $y_i^{(1)}, \dots, y_i^{(k)}$. The original support vector machine (SVM) formulation is:
\begin{equation}
\begin{array}{ll}
    \displaystyle\min_{\mathbf v,b,\bm\theta,\xi} &\frac{1}{2}\mathbf v^T\mathbf v+\frac{1}{2}\bm\theta^T\bm\theta+\kappa\displaystyle\sum_{i=1}^n\displaystyle\sum_{k=1}^{K-1}\xi_i^{(k)},\\
    \text{s.t.} &y_i^{(k)}(\langle \mathbf v,\phi(\mathbf x_i)\rangle+b-\bm\theta_k)\geq 1-\xi_i^{(k)},\\
    &\xi_i^{(k)}\geq 0, \\
    &\text{for}~i=1, \dots, n,~\text{and}~k=1, \dots, K-1,
\end{array}
\end{equation}
where $\kappa$ is a regularization constant, $\mathbf v$ are the weights, $b$ is the bias, the $\xi$ are slack variables, and $\bm\theta$ are $K-1$ thresholds.
Using the hinge loss formulation, this can be reformulated as
\begin{align}
\begin{split}
    &\displaystyle\min_{\mathbf v,b,\bm\theta,\xi} \frac{1}{2}\mathbf v^T\mathbf v+\frac{1}{2}\bm\theta^T\bm\theta\\
    &+\kappa\displaystyle\sum_{i=1}^n\displaystyle\sum_{k=1}^{K-1}\max{(0, 1-y_i^{(k)}(\langle \mathbf v,\phi(\mathbf x_i)\rangle+b-\bm\theta_k))}.
\end{split}
\end{align}
Replacing $\phi$ with a neural network $\langle \mathbf v,\phi(\mathbf x)\rangle+b$ is just a neural network with a regression head. We can also replace the hinge loss with a logistic loss which leads to a typical logistic regression objective (with a $\bm\theta_k$ subtracted inside). This is how we trained our model. Let $g(x,k)$ be the output of our trained model before the logit for data point $x$ and category $k$. The prediction is then
\begin{equation}
1+\sum_{k=1}^{K-1}\mathbbm 1_{g(x,k)>0}.
\end{equation}

\paragraph{Laplace.} For a data point $(x,y)$, we train a multi-class classifier with a softmax-head using cross entropy against the target \citep{diaz2019soft}
\begin{equation}
t_i^{(y)}=\frac{\exp^{-|y-i|}}{\sum_{k=1}^K \exp^{-|y-k|}}.
\end{equation}
Let $f(x)$ be the softmax predictions of this model. We predict the class which produces the smallest loss
\begin{equation}
\argmin_k -\sum_{i=1}^{K} f(x)_i\log(t_i^{(k)}).
\end{equation}

\paragraph{Binomial.} We start from a binomial distribution with $n$ Bernoulli trials and mean $\mu$. We first prove that if the mean has a distance of at least $2$ from both ends of the distribution, then the variance is at least one, i.e.
\begin{equation}
2\leq \mu=np\leq n-2\label{eq:meanrestri}.
\end{equation}
Now consider the variance
\begin{align}
    np(1-p) =&~ np\left(1-\frac{np}{n}\right)\\
    \geq&~np\left(1-\frac{np}{np+2}\right)\\
    =&~2\frac{np}{np+2}\\
    \geq&~1
\end{align}

We will only consider means that are natural numbers, i.e., $\mu \in \mathbb N$. Now assume that we construct a Poisson binomial distribution with mean $\mu$ but vary the variance. A Poisson binomial distribution only allows for reducing the variance. Consider $\mu$ Bernoulli trials with probability $p_1 := p+\alpha$ and $n-\mu$ Bernoulli trials with probability $p_2 := p-\frac{\mu}{n-\mu}\alpha$, then the mean will remain the same.
\begin{align}
&\mu(p+\alpha)+(n-\mu)(p-\frac{\mu}{n-\mu}\alpha)\\
= &np+\mu\alpha-(n-\mu)\frac{\mu}{n-\mu}\alpha\\
=&np=\mu
\end{align}
However, the $\alpha$ parameter allows us to modify the variance. Denote $\alpha_1:=\alpha$, and $\alpha_2:=\frac{\mu}{n-\mu}\alpha$ then the variance is
\begin{align}
    &\mu(p+\alpha_1)(1-(p+\alpha_1))\\
    +&(n-\mu)(p-\alpha_2)(1-(p-\alpha_2))\\
    =&np(1-p)-\mu\alpha_1^2-(n-\mu)\alpha_2^2\label{eq:PBdown}\\
    =&np(1-p)-\mu\alpha^2\left(1+\frac{\mu}{n-\mu}\right)\label{eq:var}
\end{align}
From eq. \ref{eq:PBdown}, it is clear that by choosing $\alpha$, we can lower the variance of the original binomial arbitrarily.

We want to represent the categories $1, \dots, K$ as soft target vectors. Since a binomial distribution with mean zero or $n$ has a zero variance, resulting in a non-soft vector, we want to fix the variance at one instead. As we previously demonstrated, this can be achieved by never considering means $0$, $1$, $n-1$, and $n$. As such, we choose a Poisson binomial with $K+3$ Bernoulli trials and represent label $y$ as mean $y+1$. Notably,
\begin{equation}
    2\leq y+1 \leq (K+3)-2.
\end{equation}
The previously considered $p$ is then $p=\frac{y+1}{K+3}$. Starting from eq. \ref{eq:var} we can calculate $\alpha$ so the variance is equal to $1$:
\begin{align}
    1=&~np(1-p)-\mu\alpha^2\left(1+\frac{\mu}{n-\mu}\right)\\
    \alpha=&~\sqrt{\frac{np(1-p)-1}{\mu(1+\frac{\mu}{n-\mu})}}\\
    =&~\sqrt{\frac{(n-\mu)^2}{n^2}-\frac{(n-\mu)}{\mu n}}\\
\end{align}
Now $n=K+3$ and $\mu=y+1$:
\begin{align}
\alpha=\alpha_1=&~\sqrt{\frac{(K-y+2)^2}{(K+3)^2}-\frac{(K-y+2)}{(y+1)(K+3)}}\\
\alpha_2=&~\sqrt{\frac{(y+1)^2}{(K+3)^2}-\frac{(y+1)}{(K-y+2)(K+3)}}
\end{align}
In practice, this means a model will have a $K+4$-dimensional softmax head. The new target for a label $y$ will be 
\begin{align}
\begin{split}
t_k^{(y)}=\sum_{i=i_0}^{i_k}&\binom{\mu}{i}p_1^i(1-p_1)^{\mu-i}\\
\cdot&\binom{n-\mu}{k-i}p_2^{k-i}(1-p_2)^{n-\mu-(k-i)},\label{eq:target}
\end{split}
\end{align}
with $i_0:=\max(0,k+\mu-n)$ and $i_k:=\min(k,\mu)$, then with this model, we follow the same prediction scheme as for Laplace, i.e., we predict the class which produces the smallest loss. In practice, it is advisable to precompute all possible binomial coefficients necessary for eq. \ref{eq:target}.

\section{Detailed Feature Extraction}

As mentioned previously, from the SM files, we extract each level sequence separately. We encode each sequence element as a 19-dimensional feature vector $x_i^{(j)}\in \mathbb R^{19}$. This vector contains the following:
\begin{itemize}
    \item the tempo in 1 feature encoded as BPM divided by $240$.
    \item an encoding of the note level in 7 features. More specifically, we encode for a note level if it is a multiple of a $\frac{1}{4}$, $\frac{1}{8}$, $\frac{1}{12}$, $\frac{1}{16}$, $\frac{1}{24}$, $\frac{1}{32}$, and other. Note that this means a note level can be a multiple of these note levels. For instance a $\frac{1}{24}$ would be encoded as $(1,1,1,0,1,0,0)$.
    \item level progress in time in 1 feature. A StepMania level has a predetermined length, so we encode the proportion of progress in this feature in time.
    \item level progress in sequence length in 1 feature. Similarly to a StepMania level's predetermined length, the sequence length of a StepMania level is predetermined. We encode the proportion of how many steps have passed over how many steps there are in total.
    \item time since the last element in 1 feature. Based on the BPM and note levels of this step and the last step, a certain amount of time has passed since the last step occurred. This time is encoded as $4$ times seconds, but it is capped at $8$.
    \item a one-hot encoding of the step direction in case of a tap in 4 features. More specifically, in case of a jump or other combination of multiple directions, the encoding will resemble a bag of steps.
    \item a one-hot encoding of the step direction in case of a hold in 4 features. The initial step of a hold is considered a tap in the encoding, all subsequent moments where the step needs to be held, we encode a bag of steps that need to be continuously pushed down. Additionally, there is an element called a roll, which means the arrow has to be tapped repeatedly at a minimum speed. Rolls are considered holds for this encoding.
\end{itemize}
This makes for a total of 19 features. Due to their rarity, we ignore all other potential features of the level sequence, like tempo changes, mines, or other effects.

\section{OR Methods Trained on StepMania-Level Difficulty Estimation Evaluated on Other Metrics}

We evaluated the experiments from Tab. \ref{tab:WAEeval}a using multiple metrics . These metrics include mean absolute error (MAE, Tab. \ref{tab:MAE}), accuracy (Tab. \ref{tab:accuracy}), true positive rate (TPR, Tab. \ref{tab:TPR}), and root mean squared error (RMSE, Tab. \ref{tab:RMSE}).

The tensor of all evaluations from Tab. \ref{tab:WAEeval}b can be found in Tab. \ref{tab:WAEevalTensor}.

\begin{table*}[tb]
    \begin{subtable}[h]{\textwidth}
    \centering
    \begin{tabular*}{\textwidth}{@{\extracolsep{\fill}}llccccc} 
 \toprule
&& ITG& fraxtil& Gpop& Gulls& Speirmix\\
\midrule
\multirow{3}{*}{non-OR}&PATTERN & 0.416 $\pm$ 0.033 & 0.614 $\pm$ 0.068 & 0.427 $\pm$ 0.026 & 0.723 $\pm$ 0.097 & 0.284 $\pm$ 0.030\\\cmidrule{2-7}
 &Classification & 0.350 $\pm$ 0.029 & 0.413 $\pm$ 0.059 &  \textbf{0.328 $\pm$ 0.026} & 0.271 $\pm$ 0.052 & 0.257 $\pm$ 0.032\\ 
 &Regression & 0.356 $\pm$ 0.027 & 0.424 $\pm$ 0.052 & 0.340 $\pm$ 0.023 & 0.283 $\pm$ 0.062 & 0.259 $\pm$ 0.032\\\midrule
 \multirow{4}{*}{OR}&NNRank & 0.357 $\pm$ 0.029 & 0.394 $\pm$ 0.048 & 0.334 $\pm$ 0.025 & 0.282 $\pm$ 0.053 & 0.254 $\pm$ 0.033\\ 
 &RED-SVM & 0.351 $\pm$ 0.026 & 0.432 $\pm$ 0.050 & 0.337 $\pm$ 0.022 &  \textbf{0.267 $\pm$ 0.053} & 0.256 $\pm$ 0.032\\ 
 &Laplace & \textbf{0.349 $\pm$ 0.028} & 0.400 $\pm$ 0.049 & 0.331 $\pm$ 0.026 & 0.274 $\pm$ 0.058 & 0.250 $\pm$ 0.033\\ 
 &Binomial & 0.351 $\pm$ 0.027 &  \textbf{0.392 $\pm$ 0.051} & 0.333 $\pm$ 0.027 & 0.270 $\pm$ 0.055 &  \textbf{0.249 $\pm$ 0.032}\\
\bottomrule \end{tabular*}
    \caption{Trained and evaluated on the same data set.}
    \end{subtable}
    \begin{subtable}[h]{\textwidth}
    \centering
    \begin{tabular*}{\textwidth}{@{\extracolsep{\fill}}llccccc} 
 \toprule
&& ITG& fraxtil& Gpop& Gulls& Speirmix\\ 
\midrule
\multirow{3}{*}{non-OR}&PATTERN& 0.577 $\pm$ 0.055& 0.535 $\pm$ 0.150& 0.648 $\pm$ 0.121& 0.498 $\pm$ 0.109& 0.544 $\pm$ 0.062\\\cmidrule{2-7}
&Classification& 0.429 $\pm$ 0.036& 0.406 $\pm$ 0.043& 0.448 $\pm$ 0.049& \textbf{0.343 $\pm$ 0.063}& 0.420 $\pm$ 0.032\\
&Regression& 0.446 $\pm$ 0.042& 0.415 $\pm$ 0.073& 0.445 $\pm$ 0.070& 0.381 $\pm$ 0.103& 0.456 $\pm$ 0.058\\\midrule
\multirow{4}{*}{OR}&NNRank& \textbf{0.412 $\pm$ 0.023}& 0.392 $\pm$ 0.042& 0.433 $\pm$ 0.054& 0.378 $\pm$ 0.082& \textbf{0.381 $\pm$ 0.064}\\
&RED-SVM& 0.424 $\pm$ 0.022& \textbf{0.387 $\pm$ 0.053}& \textbf{0.427 $\pm$ 0.065}& 0.356 $\pm$ 0.085& 0.429 $\pm$ 0.056\\
&Laplace& 0.429 $\pm$ 0.030& 0.395 $\pm$ 0.056& 0.433 $\pm$ 0.059& 0.361 $\pm$ 0.085& 0.421 $\pm$ 0.052\\
&Binomial& 0.430 $\pm$ 0.031& 0.394 $\pm$ 0.051& 0.436 $\pm$ 0.056& 0.373 $\pm$ 0.078& 0.406 $\pm$ 0.048\\
\bottomrule \end{tabular*}
    \caption{Trained on separate data sets from the one being evaluated.}
    \end{subtable}
    \caption{OR and classification models outperform feature-extraction-based PATTERN and regression models. This table shows MAE performance rounded to the third nearest digit for different models on all StepMania data sets averaged over 100 models trained on separate Monte Carlo cross-validation splits. Lower values are better. The smallest value in each column is \textbf{bold}.}
    \label{tab:MAE}
\end{table*}

\begin{table*}[tb]
    \begin{subtable}[h]{\textwidth}
    \centering
    \begin{tabular*}{\textwidth}{@{\extracolsep{\fill}}llccccc} 
 \toprule
&& ITG& fraxtil& Gpop& Gulls& Speirmix\\
\midrule
\multirow{3}{*}{non-OR}&PATTERN & 0.630 $\pm$ 0.024 & 0.519 $\pm$ 0.045 & 0.612 $\pm$ 0.019 & 0.424 $\pm$ 0.064 & 0.723 $\pm$ 0.029\\\cmidrule{2-7} 
 &Classification & 0.675 $\pm$ 0.023 & 0.629 $\pm$ 0.044 & \textbf{0.689 $\pm$ 0.019} & \textbf{0.739 $\pm$ 0.049} & 0.755 $\pm$ 0.028\\ 
 &Regression & 0.670 $\pm$ 0.022 & 0.616 $\pm$ 0.041 & 0.677 $\pm$ 0.020 & 0.725 $\pm$ 0.059 & 0.750 $\pm$ 0.027\\\midrule 
 \multirow{4}{*}{OR}&NNRank & 0.668 $\pm$ 0.024 & 0.640 $\pm$ 0.037 & 0.682 $\pm$ 0.020 & 0.725 $\pm$ 0.052 & 0.754 $\pm$ 0.030\\ 
 &RED-SVM & 0.673 $\pm$ 0.022 & 0.607 $\pm$ 0.042 & 0.679 $\pm$ 0.019 & 0.739 $\pm$ 0.051 & 0.753 $\pm$ 0.028\\ 
 &Laplace & \textbf{0.677 $\pm$ 0.023} & 0.637 $\pm$ 0.039 & 0.686 $\pm$ 0.020 & 0.737 $\pm$ 0.053 & 0.759 $\pm$ 0.028\\ 
 &Binomial & 0.676 $\pm$ 0.021 & \textbf{0.644 $\pm$ 0.040} & 0.685 $\pm$ 0.020 & 0.741 $\pm$ 0.051 & \textbf{0.760 $\pm$ 0.027}\\
\bottomrule \end{tabular*}
    \caption{Trained and evaluated on the same data set.}
    \end{subtable}
    \begin{subtable}[h]{\textwidth}
    \centering
    \begin{tabular*}{\textwidth}{@{\extracolsep{\fill}}llccccc} 
 \toprule
&& ITG& fraxtil& Gpop& Gulls& Speirmix\\ 
\midrule
\multirow{3}{*}{non-OR}&PATTERN& 0.538 $\pm$ 0.021& 0.556 $\pm$ 0.069& 0.492 $\pm$ 0.053& 0.549 $\pm$ 0.068& 0.513 $\pm$ 0.041\\\cmidrule{2-7}
&Classification& 0.629 $\pm$ 0.017& 0.636 $\pm$ 0.029& 0.604 $\pm$ 0.037& \textbf{0.670 $\pm$ 0.057}& 0.605 $\pm$ 0.032\\
&Regression& 0.605 $\pm$ 0.023& 0.628 $\pm$ 0.056& 0.596 $\pm$ 0.053& 0.639 $\pm$ 0.091& 0.571 $\pm$ 0.050\\\midrule
\multirow{4}{*}{OR}&NNRank& \textbf{0.632 $\pm$ 0.014}& 0.639 $\pm$ 0.036& 0.606 $\pm$ 0.045& 0.642 $\pm$ 0.074& \textbf{0.636 $\pm$ 0.060}\\
&RED-SVM& 0.621 $\pm$ 0.017& \textbf{0.645 $\pm$ 0.044}& 0.608 $\pm$ 0.049& 0.658 $\pm$ 0.079& 0.595 $\pm$ 0.050\\
&Laplace& 0.621 $\pm$ 0.012& 0.637 $\pm$ 0.047& \textbf{0.609 $\pm$ 0.047}& 0.657 $\pm$ 0.077& 0.602 $\pm$ 0.049\\
&Binomial& 0.623 $\pm$ 0.012& 0.638 $\pm$ 0.043& 0.608 $\pm$ 0.046& 0.648 $\pm$ 0.071& 0.616 $\pm$ 0.045\\
\bottomrule \end{tabular*}
    \caption{Trained on separate data sets from the one being evaluated.}
    \end{subtable}
    \caption{OR and classification models outperform feature-extraction-based PATTERN and regression models. This table shows accuracy performance rounded to the third nearest digit for different models on all StepMania data sets averaged over 100 models trained on separate Monte Carlo cross-validation splits. Lower values are better. The smallest value in each column is \textbf{bold}.}
    \label{tab:accuracy}
\end{table*}


\begin{table*}[tb]
    \begin{subtable}[h]{\textwidth}
    \centering
    \begin{tabular*}{\textwidth}{@{\extracolsep{\fill}}llccccc} 
 \toprule
&& ITG& fraxtil& Gpop& Gulls& Speirmix\\
\midrule
\multirow{3}{*}{non-OR}&PATTERN & 0.592 $\pm$ 0.034 & 0.434 $\pm$ 0.058 & 0.587 $\pm$ 0.021 & 0.440 $\pm$ 0.071 & 0.711 $\pm$ 0.034\\\cmidrule{2-7} 
 &Classification & 0.653 $\pm$ 0.027 & 0.578 $\pm$ 0.060 & \textbf{0.676 $\pm$ 0.021} & 0.740 $\pm$ 0.061 & 0.742 $\pm$ 0.038\\ 
 &Regression & 0.648 $\pm$ 0.027 & 0.555 $\pm$ 0.047 & 0.662 $\pm$ 0.022 & 0.722 $\pm$ 0.063 & 0.737 $\pm$ 0.032\\\midrule 
 \multirow{4}{*}{OR}&NNRank & 0.651 $\pm$ 0.027 & 0.592 $\pm$ 0.049 & 0.673 $\pm$ 0.021 & 0.735 $\pm$ 0.063 & 0.746 $\pm$ 0.037\\ 
 &RED-SVM & 0.656 $\pm$ 0.027 & 0.555 $\pm$ 0.047 & 0.668 $\pm$ 0.021 & 0.740 $\pm$ 0.057 & 0.747 $\pm$ 0.033\\ 
 &Laplace & 0.658 $\pm$ 0.025 & 0.586 $\pm$ 0.055 & 0.675 $\pm$ 0.020 & 0.739 $\pm$ 0.062 & 0.750 $\pm$ 0.036\\ 
 &Binomial & \textbf{0.659 $\pm$ 0.025} & \textbf{0.595 $\pm$ 0.056} & 0.674 $\pm$ 0.021 & \textbf{0.746 $\pm$ 0.062} & \textbf{0.750 $\pm$ 0.036}\\ 
\bottomrule \end{tabular*}
    \caption{Trained and evaluated on the same data set.}
    \end{subtable}
    \begin{subtable}[h]{\textwidth}
    \centering
    \begin{tabular*}{\textwidth}{@{\extracolsep{\fill}}llccccc} 
 \toprule
&& ITG& fraxtil& Gpop& Gulls& Speirmix\\ 
\midrule
\multirow{3}{*}{non-OR}&PATTERN& 0.462 $\pm$ 0.055& 0.491 $\pm$ 0.082& 0.450 $\pm$ 0.071& 0.517 $\pm$ 0.084& 0.455 $\pm$ 0.070\\\cmidrule{2-7}
&Classification& \textbf{0.596 $\pm$ 0.034}& 0.571 $\pm$ 0.042& 0.570 $\pm$ 0.040& \textbf{0.643 $\pm$ 0.076}& 0.553 $\pm$ 0.040\\
&Regression& 0.564 $\pm$ 0.030& 0.573 $\pm$ 0.068& 0.563 $\pm$ 0.061& 0.614 $\pm$ 0.104& 0.523 $\pm$ 0.073\\\midrule
\multirow{4}{*}{OR}&NNRank& 0.593 $\pm$ 0.026& 0.575 $\pm$ 0.057& 0.574 $\pm$ 0.051& 0.624 $\pm$ 0.091& \textbf{0.590 $\pm$ 0.070}\\
&RED-SVM& 0.578 $\pm$ 0.031& \textbf{0.585 $\pm$ 0.064}& 0.574 $\pm$ 0.054& 0.632 $\pm$ 0.092& 0.534 $\pm$ 0.070\\
&Laplace& 0.581 $\pm$ 0.026& 0.580 $\pm$ 0.062& \textbf{0.578 $\pm$ 0.054}& 0.637 $\pm$ 0.093& 0.551 $\pm$ 0.061\\
&Binomial& 0.582 $\pm$ 0.025& 0.580 $\pm$ 0.060& 0.576 $\pm$ 0.053& 0.632 $\pm$ 0.089& 0.567 $\pm$ 0.059\\
\bottomrule \end{tabular*}
    \caption{Trained on separate data sets from the one being evaluated.}
    \end{subtable}
    \caption{OR and classification models outperform feature-extraction-based PATTERN and regression models. Binomial performs especially well. This table shows TPR performance rounded to the third nearest digit for different models on all StepMania data sets averaged over 100 models trained on separate Monte Carlo cross-validation splits. Lower values are better. The smallest value in each column is \textbf{bold}.}
    \label{tab:TPR}
\end{table*}


\begin{table*}[tb]
    \begin{subtable}[h]{\textwidth}
    \centering
    \begin{tabular*}{\textwidth}{@{\extracolsep{\fill}}llccccc} 
 \toprule
&& ITG& fraxtil& Gpop& Gulls& Speirmix\\
\midrule
\multirow{3}{*}{non-OR}&PATTERN & 0.745 $\pm$ 0.092 & 0.964 $\pm$ 0.086 & 0.715 $\pm$ 0.032 & 1.028 $\pm$ 0.113 & 0.544 $\pm$ 0.032\\\cmidrule{2-7} 
 &Classification & 0.642 $\pm$ 0.049 & 0.708 $\pm$ 0.074 & 0.610 $\pm$ 0.070 & 0.537 $\pm$ 0.060 & 0.540 $\pm$ 0.060\\ 
 &Regression & 0.646 $\pm$ 0.036 & 0.709 $\pm$ 0.060 & 0.616 $\pm$ 0.038 & 0.543 $\pm$ 0.066 & 0.530 $\pm$ 0.050\\\midrule 
 \multirow{4}{*}{OR}&NNRank & 0.641 $\pm$ 0.037 & \textbf{0.678 $\pm$ 0.058} & 0.612 $\pm$ 0.052 & 0.540 $\pm$ 0.055 & 0.524 $\pm$ 0.050\\ 
 &RED-SVM & \textbf{0.637 $\pm$ 0.035} & 0.712 $\pm$ 0.056 & \textbf{0.610 $\pm$ 0.038} & \textbf{0.524 $\pm$ 0.058} & 0.527 $\pm$ 0.050\\ 
 &Laplace & 0.637 $\pm$ 0.037 & 0.688 $\pm$ 0.059 & 0.612 $\pm$ 0.065 & 0.539 $\pm$ 0.069 & 0.528 $\pm$ 0.066\\ 
 &Binomial & 0.642 $\pm$ 0.039 & 0.682 $\pm$ 0.064 & 0.617 $\pm$ 0.068 & 0.537 $\pm$ 0.065 & \textbf{0.523 $\pm$ 0.060}\\ 
\bottomrule \end{tabular*}
    \caption{Trained and evaluated on the same data set.}
    \end{subtable}
    \begin{subtable}[h]{\textwidth}
    \centering
    \begin{tabular*}{\textwidth}{@{\extracolsep{\fill}}llccccc} 
 \toprule
&& ITG& fraxtil& Gpop& Gulls& Speirmix\\ 
\midrule
\multirow{3}{*}{non-OR}&PATTERN& 0.949 $\pm$ 0.073& 0.851 $\pm$ 0.199& 0.993 $\pm$ 0.152& 0.767 $\pm$ 0.124& 0.816 $\pm$ 0.069\\\cmidrule{2-7}
&Classification& 0.788 $\pm$ 0.104& 0.714 $\pm$ 0.070& 0.789 $\pm$ 0.088& \textbf{0.607 $\pm$ 0.060}& 0.699 $\pm$ 0.026\\
&Regression& 0.751 $\pm$ 0.056& 0.702 $\pm$ 0.074& 0.733 $\pm$ 0.075& 0.644 $\pm$ 0.092& 0.722 $\pm$ 0.054\\\midrule
\multirow{4}{*}{OR}&NNRank& \textbf{0.726 $\pm$ 0.046}& 0.674 $\pm$ 0.039& 0.730 $\pm$ 0.050& 0.644 $\pm$ 0.075& \textbf{0.653 $\pm$ 0.054}\\
&RED-SVM& 0.730 $\pm$ 0.037& \textbf{0.673 $\pm$ 0.051}& \textbf{0.715 $\pm$ 0.067}& 0.616 $\pm$ 0.076& 0.701 $\pm$ 0.048\\
&Laplace& 0.760 $\pm$ 0.081& 0.679 $\pm$ 0.055& 0.746 $\pm$ 0.069& 0.627 $\pm$ 0.077& 0.700 $\pm$ 0.046\\
&Binomial& 0.764 $\pm$ 0.088& 0.678 $\pm$ 0.051& 0.750 $\pm$ 0.065& 0.641 $\pm$ 0.070& 0.683 $\pm$ 0.042\\
\bottomrule \end{tabular*}
    \caption{Trained on separate data sets from the one being evaluated.}
    \end{subtable}
    \caption{OR and classification models outperform feature-extraction-based PATTERN and regression models. This table shows RMSE performance rounded to the third nearest digit for different models on all StepMania data sets averaged over 100 models trained on separate Monte Carlo cross-validation splits. Lower values are better. The smallest value in each column is \textbf{bold}.}
    \label{tab:RMSE}
\end{table*}

\begin{table*}
    \centering
    \begin{tabular*}{\textwidth}{@{\extracolsep{\fill}}llccccc} 
    \toprule
    &Trained on & ITG & fraxtil & Gpop & Gulls & Speirmix\\
    \midrule
    \multirow{5}{*}{PATTERN}& ITG& -& 0.508& 0.544& 0.426& 0.477\\
& fraxtil& 0.746& -& 0.809& 0.723& 0.728\\
& Gpop& 0.633& 0.587& -& 0.501& 0.724\\
& Gulls& 0.928& 0.879& 0.903& -& 0.749\\
& Speirmix& 0.561& 0.498& 0.613& 0.544& -\\
\midrule
\multirow{5}{*}{Classification}& ITG& -& 0.414& 0.401& 0.324& 0.420\\
& fraxtil& 0.491& -& 0.495& 0.422& 0.527\\
& Gpop& 0.381& 0.445& -& 0.293& 0.541\\
& Gulls& 0.520& 0.509& 0.518& -& 0.551\\
& Speirmix& 0.451& 0.541& 0.536& 0.476& -\\
\midrule
\multirow{5}{*}{Regression}& ITG& -& 0.393& 0.378& 0.322& 0.413\\
& fraxtil& 0.507& -& 0.481& 0.384& 0.640\\
& Gpop& 0.427& 0.429& -& 0.347& 0.575\\
& Gulls& 0.535& 0.480& 0.487& -& 0.619\\
& Speirmix& 0.463& 0.574& 0.573& 0.590& -\\
\midrule
\multirow{5}{*}{NNRank}& ITG& -& 0.412& 0.413& 0.410& 0.351\\
& fraxtil& 0.489& -& 0.450& 0.348& 0.563\\
& Gpop& 0.400& 0.431& -& 0.305& 0.504\\
& Gulls& 0.498& 0.443& 0.452& -& 0.471\\
& Speirmix& 0.434& 0.557& 0.562& 0.550& -\\
\midrule
\multirow{5}{*}{RED-SVM}& ITG& -& 0.392& 0.381& 0.316& 0.410\\
& fraxtil& 0.497& -& 0.460& 0.372& 0.630\\
& Gpop& 0.408& 0.418& -& 0.321& 0.563\\
& Gulls& 0.508& 0.440& 0.465& -& 0.581\\
& Speirmix& 0.448& 0.560& 0.556& 0.539& -\\
\midrule
\multirow{5}{*}{Laplace}& ITG& -& 0.394& 0.383& 0.344& 0.394\\
& fraxtil& 0.496& -& 0.462& 0.347& 0.592\\
& Gpop& 0.415& 0.413& -& 0.312& 0.531\\
& Gulls& 0.545& 0.464& 0.478& -& 0.598\\
& Speirmix& 0.444& 0.559& 0.551& 0.548& -\\
\midrule
\multirow{5}{*}{Binomial}& ITG& -& 0.398& 0.390& 0.356& 0.378\\
& fraxtil& 0.495& -& 0.470& 0.363& 0.544\\
& Gpop& 0.416& 0.418& -& 0.315& 0.522\\
& Gulls& 0.545& 0.453& 0.473& -& 0.546\\
& Speirmix& 0.444& 0.558& 0.552& 0.546& -\\
    \bottomrule
    \end{tabular*}
    \caption{Generalization from one data set to another depends on how well the labels align. Training on ITG appears to lead to a rather good generalization to other data sets. This table is a visualization of the tensor containing all individual results used for the computation of Tab. \ref{tab:WAEeval}b}
    \label{tab:WAEevalTensor}
\end{table*}

\begin{figure*}[!tb]
\centering
\begin{subfigure}[b]{.33\textwidth}
    \includegraphics[width=\linewidth]{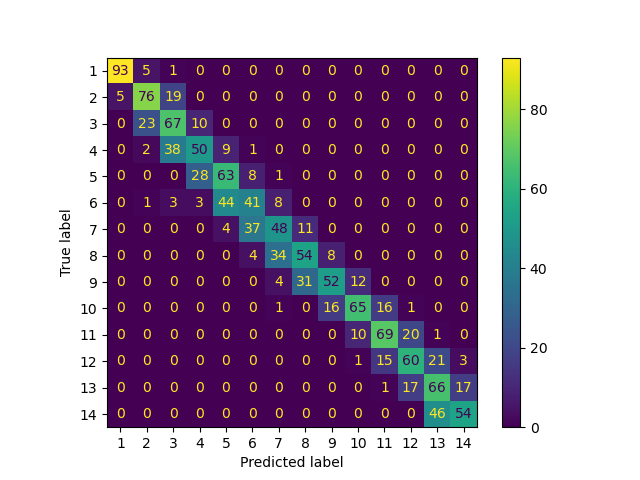}
    \caption{Trained on the Gpop with Laplace and evaluated on ITG. They starts aligned but tend towards the lower class}
\end{subfigure}
\begin{subfigure}[b]{.33\textwidth}
    \includegraphics[width=\linewidth]{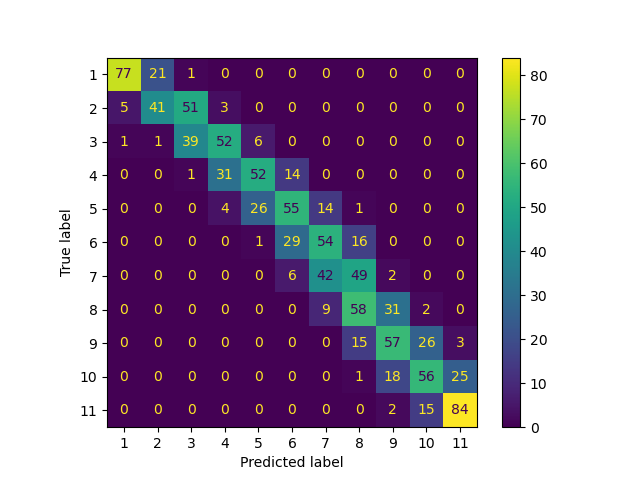}
    \caption{Trained on Speirmix with Regression and evaluated on Gpop. Strong tendency towards the next higher class.}
\end{subfigure}
\begin{subfigure}[b]{.33\textwidth}
    \includegraphics[width=\linewidth]{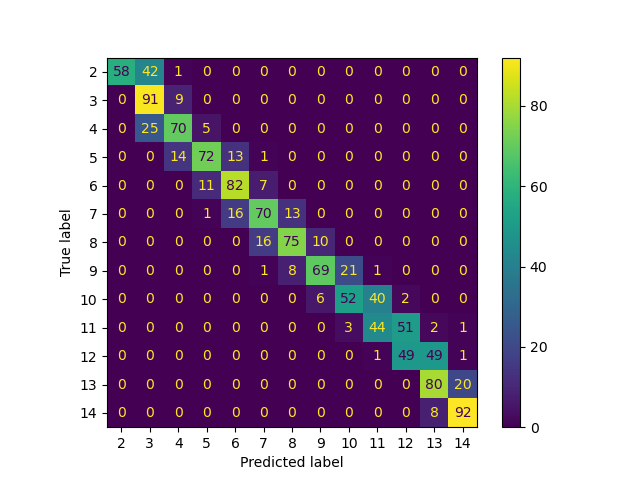}
    \caption{Trained on Gpop with Binomial and evaluated on Gulls. The two data sets are mostly aligned}
\end{subfigure}
\caption{The average and category normalized confusion matrix of our model trained various data sets with the various models evaluated on various data sets.}
\label{fig:confusion-others}
\end{figure*}

We also visualized more confusion matrices across data sets. Find them in Fig. 
\ref{fig:confusion-others}.

\section{Alternative Agreement Evaluations}

Tab. \ref{tab:agreement2} shows a similar evaluation to Tab. \ref{tab:agreement}. The difference is that in Tab. \ref{tab:agreement2}, no pairs are disregarded during evaluation, i.e., even if a model and the original labels disagree and one of the two claims equality of the pair, this pair will be considered classified incorrectly. This leads to a worse generalization than Tab. \ref{tab:agreement}. The generalization error moves from $\approx 0.002$ to $\approx 0.015$. This error is caused by the equality in difficulty ranking indicating an unknown ranking, assuming two elements cannot be equal, which is a reasonable assumption for StepMania levels.

\begin{table*}[tb]
    \begin{subtable}[h]{\textwidth}
    \centering
    \begin{tabular*}{\textwidth}{@{\extracolsep{\fill}}llccccc}
    \toprule
    && ITG & fraxtil & Gpop & Gulls & Speirmix\\ 
    \multirow{3}{*}{non-OR}&PATTERN & 0.902 $\pm$ 0.006 & 0.904 $\pm$ 0.009 & 0.909 $\pm$ 0.004 & 0.848 $\pm$ 0.016 & 0.911 $\pm$ 0.008\\
    \cmidrule{2-7}
    &Classification & 0.912 $\pm$ 0.006 & 0.925 $\pm$ 0.009 & \textbf{0.924 $\pm$ 0.005} & 0.919 $\pm$ 0.016 & 0.917 $\pm$ 0.009\\ 
    &Regression & 0.911 $\pm$ 0.006 & 0.924 $\pm$ 0.008 & 0.922 $\pm$ 0.004 & 0.918 $\pm$ 0.016 & 0.916 $\pm$ 0.009\\ 
    \midrule
    \multirow{4}{*}{OR}&NNRank & 0.911 $\pm$ 0.006 & 0.927 $\pm$ 0.008 & 0.923 $\pm$ 0.004 & 0.919 $\pm$ 0.015 & \textbf{0.919 $\pm$ 0.009}\\ 
    &RED-SVM & 0.911 $\pm$ 0.006 & 0.920 $\pm$ 0.009 & 0.922 $\pm$ 0.004 & \textbf{0.921 $\pm$ 0.017} & 0.916 $\pm$ 0.009\\ 
    &Laplace & \textbf{0.912 $\pm$ 0.006} & 0.927 $\pm$ 0.008 & 0.923 $\pm$ 0.005 & 0.918 $\pm$ 0.018 & 0.918 $\pm$ 0.009\\ 
    &Binomial & 0.912 $\pm$ 0.006 & \textbf{0.928 $\pm$ 0.008} & 0.923 $\pm$ 0.005 & 0.920 $\pm$ 0.017 & 0.918 $\pm$ 0.009\\ 
    \bottomrule
    \end{tabular*}
    \caption{Trained and evaluated on the same data set.}
    \end{subtable}
    \begin{subtable}[h]{\textwidth}
    \centering
    \begin{tabular*}{\textwidth}{@{\extracolsep{\fill}}llccccc}
    \toprule
    && ITG & fraxtil & Gpop & Gulls & Speirmix\\
    \midrule
    \multirow{3}{*}{non-OR}&PATTERN& 0.875 $\pm$ 0.017& 0.896 $\pm$ 0.027& 0.873 $\pm$ 0.019& 0.877 $\pm$ 0.014& 0.873 $\pm$ 0.026\\
    \cmidrule{2-7}
    &Classification& 0.898 $\pm$ 0.012& 0.916 $\pm$ 0.010& 0.905 $\pm$ 0.006& 0.903 $\pm$ 0.015& 0.891 $\pm$ 0.005\\
    &Regression& 0.897 $\pm$ 0.012& 0.918 $\pm$ 0.011& 0.908 $\pm$ 0.008& 0.902 $\pm$ 0.018& 0.892 $\pm$ 0.006\\
    \midrule
    \multirow{4}{*}{OR}&NNRank& \textbf{0.900 $\pm$ 0.010}& \textbf{0.921 $\pm$ 0.006}& \textbf{0.910 $\pm$ 0.005}& 0.903 $\pm$ 0.017& \textbf{0.898 $\pm$ 0.005}\\
    &RED-SVM& 0.898 $\pm$ 0.009& 0.920 $\pm$ 0.007& 0.908 $\pm$ 0.007& \textbf{0.905 $\pm$ 0.017}& 0.891 $\pm$ 0.006\\
    &Laplace& 0.898 $\pm$ 0.012& 0.919 $\pm$ 0.009& 0.908 $\pm$ 0.007& 0.903 $\pm$ 0.018& 0.893 $\pm$ 0.006\\
    &Binomial& 0.898 $\pm$ 0.012& 0.919 $\pm$ 0.009& 0.908 $\pm$ 0.006& 0.901 $\pm$ 0.016& 0.894 $\pm$ 0.006\\
    \bottomrule
    \end{tabular*}
    \caption{Trained on separate data sets from the one being evaluated.}
    \end{subtable}
    \caption{Generalization is close to perfect when considering StepMania difficulty prediction as a ranking problem. This table shows agreement of different models on all StepMania data sets averaged over 100 models trained on separate Monte Carlo cross-validation splits. Higher values are better. The largest value in each column is \textbf{bold}. In contrast to Tab. \ref{tab:agreement}, this table contains the full accuracy considering equality a separate label.}
    \label{tab:agreement2}
\end{table*}

\section{Alternative User Feedback Evaluations}

\paragraph{Metric.} The metric used for Tab. \ref{tab:user-test} and Tab. \ref{tab:user-test2} is an accuracy based on concordance. Let $D$ be a set of data points $x$ and $D':=\{(x,x')|x,x'\in D\}$ be the set of data point pairs. Each data point pair will have a concordance $r_{x,x'}$ with an ordering $rank(x)>rank(x')$, i.e., $Pr[rank(x)>rank(x')]=r_{x,x'}$. The accuracy $\text{Acc}^{D,r}_f$ of a model $f$, or agreement, on this concordance would then be
\begin{equation}
    \text{Acc}^r_D(f) = \sum_{(x,x')\in D'} r_{x,x'}\mathbbm 1_{f(x,x')=[[rank(x) > rank(x')]]}
\end{equation}
This is akin to classification with uncertain labels.

Tab. \ref{tab:user-test2} shows a similar evaluation to Tab. \ref{tab:user-test}. In Tab. \ref{tab:user-test2}, every model is evaluated on the same data set it was trained on. For a model, pairs containing data points on which the model was trained are excluded from this evaluation to avoid biased results.

\begin{table}[tb]
    \centering
    \resizebox{\linewidth}{!}{
     \begin{tabular}{lrrrrr} 
     \toprule
    & ITG& fraxtil & Gpop & Gulls & Speirmix\\
    \midrule
    Original&  \textbf{0.675} & 0.470 & 0.257 & 0.607 & 0.388\\\midrule
    PATTERN & 0.533 & 0.625 & 0.465 & 0.304& 0.572 \\ 
 Classification & 0.353 & 0.715& \textbf{0.774}& 0.646& 0.688\\ \midrule
 Regression & 0.517 & 0.734 & 0.755& 0.566& 0.701\\ 
 NNRank & 0.41 & \textbf{0.736} & 0.74 & 0.59 & 0.715\\ 
 RED-SVM & 0.511& 0.729& 0.734 & \textbf{0.691}& \textbf{0.722}\\ 
 Laplace & 0.466& 0.723 & 0.738& 0.674 & 0.704\\ 
 Binomial & 0.459 & 0.717 & 0.745 & 0.59 & 0.681\\ 
    \bottomrule \end{tabular}}
    \caption{Most models improve upon the original labels of each data set, except for ITG. This table shows the agreement with user evaluations on various data sets of models trained on the same data set, in contrast to Tab. \ref{tab:user-test}}
    \label{tab:user-test2}
\end{table}

\section{Other Related Work}
We would also like to mention related work from a journal that proclaims itself to be about joke research, namely, the Special Interest Group on Harry Quimby Bovik (SIGBOVIK) from the association for computational heresy.
\citet{blum2016turn, blum2017cross, blum2019bracket}
~presents work on ITG (considered synonymous with StepMania) analyzing different StepMania-level elements that have become more standard over time. These StepMania level elements affect the difficulty perception of levels over time as new approaches to overcome challenges become commonplace. One such example, perhaps the latest, would be the adoption of so-called brackets. Doing a ``bracket'' means stepping onto two adjacent keys simultaneously instead of jumping to push each of these keys separately with separate feet. This directly affects difficulty, as it allows for pressing more than two keys at once, which in the past required the additional use of hands and uncomfortable body positioning. Be aware that this related work is, in large part, ironic.


\end{document}